\documentclass[article]{llncs}
\usepackage{lipsum}
\usepackage{amsmath}
\usepackage{amssymb}
\usepackage[boxruled,vlined,linesnumbered]{algorithm2e}
\usepackage{graphics}
\usepackage{epsfig}
\usepackage{color}

\DeclareMathOperator*{\argmax}{\arg\max}
\newcommand\set[1]{\mathcal{#1}}

\renewcommand\footnotemark{}

\title{Using Monte Carlo Search With Data Aggregation to Improve Robot
  Soccer Policies}
  
\author{Francesco Riccio\inst{1}*\thanks{* These two authors have contributed equally to the work.} \and Roberto Capobianco\inst{1}* \and Daniele Nardi\inst{1}}

\institute{Department of Computer, Control and Management Engineering
  ``Antonio Ruberti'', Sapienza University of Rome, Rome, Italy\\
  \email{\{riccio, capobianco, nardi\}@dis.uniroma1.it}}

\begin{document}
\maketitle

\begin{abstract}
  RoboCup soccer competitions are considered among the most
  challenging multi-robot adversarial environments, due to their high
  dynamism and the partial observability of the environment. In this
  paper we introduce a method based on a combination of Monte Carlo
  search and data aggregation (MCSDA) to adapt discrete-action soccer
  policies for a defender robot to the strategy of the opponent
  team. By exploiting a simple representation of the domain, a
  supervised learning algorithm is trained over an initial collection
  of data consisting of several simulations of human expert
  policies. Monte Carlo policy rollouts are then generated and
  aggregated to previous data to improve the learned policy over
  multiple epochs and games. The proposed approach has been
  extensively tested both on a soccer-dedicated simulator and on real
  robots. Using this method, our learning robot soccer team achieves
  an improvement in ball interceptions, as well as a reduction in the
  number of opponents' goals. Together with a better performance, an
  overall more efficient positioning of the whole team within the
  field is achieved.
  
\end{abstract}

\section{INTRODUCTION}
\label{sec:introduction}

Machine learning methods have been increasingly used in robotics to
deal with uncertain and unstructured environments. In such scenarios,
directly learning from data a (sub-)optimal set of parameters to
generate robot behaviors is often more robust than hard coding them
from prior knowledge. However, the variety of the problems and the
lack of big amount of data still refrain researchers from the
application of standard learning approaches to challenging domains
such as RoboCup soccer competitions~\cite{Kitano1997}. Here, in fact,
manifold problems must be faced by a multi-robot system, such as
coordination and decision making under partial observability of an
adversarial and dynamic environment.

RoboCup soccer teams typically tackle competitions by deploying static
behaviors for their robots. Here, each programmed agent executes a
single policy that takes into account the state of the robot teammates,
but does not change at run-time. However, during the game the partial
observable environment discloses previously unavailable information,
such as the strategy of the opponent team. On the one hand, predefined
behavioral protocols cannot handle the newly available
knowledge. Hence, the use of a learning approach to update each
agent's current policy would be beneficial to the team performance. On
the other hand, such information only consists of small portions of
new data, that cannot be used without exploiting the structure of the
domain and adequate machine learning methods.

\begin{figure}[!t]
\centering
\includegraphics[width=\textwidth]{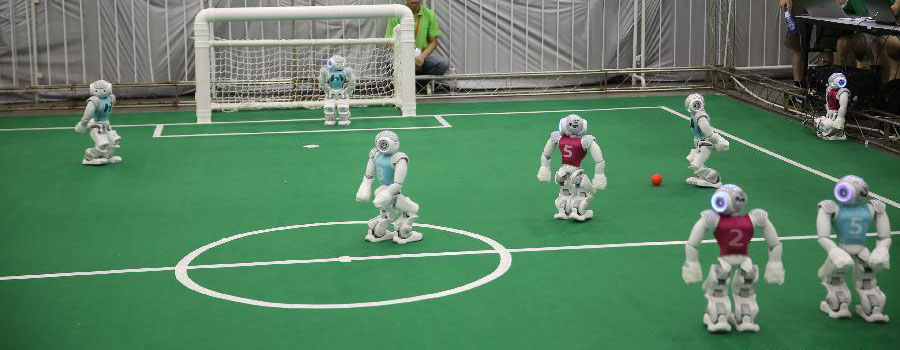}
\caption{MCSDA generates effective robot policies in a highly dynamic
  environment.}
\label{fig:intro}
\end{figure}

In this paper, we specifically consider the setup proposed by the
RoboCup Standard Platform League (SPL), where NAO robots compete in a
5-vs-5 soccer game (see \figurename~\ref{fig:intro}). Our goal
consists in generating a robot defender policy that adapts to the
strategy of the opponent team. Such strategy is not known and, hence,
makes the world dynamics unknown or difficult to model. The policy
that we generate is composed of a discrete and limited set of actions,
and it is at first instantiated to imitate an initial dataset of human
(expert) deployed behaviors. To this end, we introduce Monte Carlo
Search with Data Aggregation (MCSDA). Our algorithm uses a standard
classifier to imitate expert actions given the current observation of
a simplified representation of the game domain, modeled by the
position and velocity of the ball in the field, as well as the player
position. Since the classifier is trained over the distribution of
observations and expert actions from multiple games, frequent patterns
and main game areas can be exploited by the learned policy. Such
policy is then improved by aggregating~\cite{Ross2011} the initial
dataset with policy rollouts collected using simple Monte Carlo
search~\cite{Tesauro1996}. While our algorithm strictly relates to
state-of-the-art methods for reinforcement learning with unknown
system dynamics~\cite{Ross2014} and recent applications to games like
Go~\cite{Silver2016}, the main novelty of this paper consists in the
combination of these techniques, that allows to achieve good results
on a partially observable, high dynamic robotic context. With this
paper, in fact, we aim at showing that (1) the use of data aggregation
together with Monte Carlo search is practical, effectively improves
the learner's policy and preserves good properties, and (2) by
adopting a simplified representation of the domain a good policy
improvement can be obtained on complex and challenging robotic
scenarios. The obtained results show improvements in the overall team
performance, where the percentage of recovered ball and the number of
won games increase with the number of MCSDA iterations.

The reminder of this paper is organized as
follows. Section~\ref{sec:related-work} provides an overview on the
literature about policy learning and improvement, as well as strategy
adaptation in the RoboCup context; Section~\ref{sec:methodology}
describes in detail the proposed approach introducing the MCSDA
algorithm (Section~\ref{sec:monte-carlo}). Finally,
Section~\ref{sec:exper-eval} describes the robot platform and the
experimental setup together with the obtained results, while
Section~\ref{sec:conclusions} concludes the paper with final remarks
and future work.

\section{RELATED WORK}
\label{sec:related-work}

Policy learning is a very active area of research, due to its
complexity and practical relevance. Reinforcement learning, Monte
Carlo methods and imitation learning have been successfully applied in
several contexts and domains. For example, in robotics Kober and
Peters~\cite{Kober2009} use episodic reinforcement learning in order
to improve motor primitives learned by imitation for a Ball-in-a-Cup
task. Kormushev et al.~\cite{Kormushev2010}, instead, encode
movements with and extension of Dynamic Movement
Primitives~\cite{Ijspeert2001} initialized from
imitation. Reinforcement learning is then used to learn the optimal
parameters of the policy, thus improving the obtained
performance. Differently, Ross et al.~\cite{Ross2011} propose a
meta-algorithm for imitation learning (\textsc{DAgger}), which learns
a stationary deterministic policy that is guaranteed to perform well
under its induced distribution of states. Their method, which strictly
relates to a no-regret online learning approach, is then applied to
learn some policies that can steer a car in a 3D racing game and can
play Super Mario Bros., given input image features and corresponding
expert demonstrations. The idea of applying policy learning on
video-games has been recently used also by Mnih et
  al.~\cite{Mnih2015}, that present a deep agent (deep Q-network),
that can use reinforcement learning to generate policies directly from
high-dimensional sensory inputs. The authors test their algorithm on
classic Atari 2600 games, achieving a level comparable to that of a
professional human player across a set of 49 games. Similarly, Silver
et al.~\cite{Silver2016} use deep ``value networks'' and
``policy networks'' to respectively evaluate board positions and
select moves for the challenging game of Go. These neural networks are
trained by a combination of supervised learning from human expert
games, reinforcement learning and Monte Carlo tree search. The
resulting program showed to be able to beat human Go champions and to
achieve a performance beyond any previous expectation.

Building on the idea of adopting a combination of techniques similar
to~\cite{Silver2016}, our work mostly relates to the
\textsc{AggreVate} and NRPI algorithms by Ross and
Bagnell~\cite{Ross2014}. The former leverages cost-to-go information
-- in addition to correct demonstration -- and data aggregation; the
latter extends the idea of no-regret learners to Approximate Policy
Iteration variants for reinforcement learning. However, differently
from previous work our algorithm (MCSDA) uses shorter Monte Carlo
roll-outs to evaluate policy improvements. By avoiding to always
estimate the full cost-to-go of the policy MCSDA is more
practical -- and usable in robotics.  Additionally, as explained in
Section~\ref{sec:methodology}, the policy generated by our algorithm
can be seen as a combination of expert and learned policies, allowing
us to directly leverage results from \cite{Chang2015}.

\subsection*{Policy Adaptation in RoboCup}
\label{sec:policy-learn-roboc}

Policy classification and adaptation to the strategy of the opponent
team is not a new idea in RoboCup competitions. For example, Han and
Veloso~\cite{Han2000} propose to employ Hidden Markov Models to detect
opponents' behaviors, represented as game states. The authors first
characterize the game state in terms of ``behavioral-relevant state
features'' and then show how a cascade of HMMs is able to recognize
different pre-defined robot behaviors. This idea has been further
developed by Riley et al.~\cite{Riley2000}, who propose a
classification method for the opponents' behavior in a simulated
environment. The authors first enable their agents to observe and
classify the actions of the adversaries, and then to accordingly adapt
their policy. More recently, Trevizan and Veloso~\cite{Trevizan2010}
also address the problem of classifying opponents, and their
strategies with respect to a set of behavioral
components. Specifically, they are able to generalize and classify
unknown opponents as combination of known ones. Yasui et
al.~\cite{Yasui2013} introduce a ``dissimilarity function'' to
categorize opponent strategies via cluster analysis. The authors
improve their team performance by analyzing logged data of previous
matches and showing that team attacking strategies can be recognized
and correctly classified. Finally, Biswas et al.~\cite{Biswas2014}
propose an opponent aware defensive strategies. In particular, once
the state of the opponents is received, the robotic systems categorize
the attacking robot as first and second level threats. Accordingly,
the team displaces a variable number of defenders in order to prevent
the opponent team to score.

It is worth remarking that all the aforementioned methods propose
effective solutions to the problem of decision making in presence of
adversaries. However, differently from our application of MCSDA on the
RoboCup scenario -- that uses only a small portion of the game-state
and operates under unknown system dynamics, they operate in controlled
environments, where full information is available. Additionally, while
the MCSDA algorithm can be separately applied on each agent and
automatically accounts for uncertainty, the described systems are
usually centralized and do not consider uncertain outcomes. For these
reasons, we consider our approach a valuable contribution also to the
robotic and RoboCup community, where partially observable and highly
dynamic scenarios need to be addressed.

\section{APPROACH}
\label{sec:methodology}

The generation of our adaptive policy relies on standard machine
learning methods. First, a classifier is used to imitate a sub-optimal
expert policy and accordingly choose an action, given the current
observation of the game domain. Then, the learned policy is improved
by aggregating, in an online learning fashion, previous data with
Monte Carlo policy rollouts. Throughout the learning process, the
domain representation is simplified and it is reduced to the essential
game elements -- the position and velocity of the ball in the field.

\subsection{Preliminaries}
\label{sec:preliminaries}

We present our learning problem using the Markov Decision Process
notation, where $S$ and $A$ respectively represent a discrete set of
states and actions, and $R(s)$ is the immediate reward obtained for
being in state $s \in S$. $R$ is assumed to be bounded in $[0, 1]$. In
our learning setting not only we observe the reward function $R$, but
also demonstrations of a sub-optimal policy $\pi^*$ that aims at
maximizing $R$ and induces a state distribution
$d_{\pi^{*}}$. Additionally, we assume the dynamics of the world to be
unknown or to be accessible only through samples, due to its
complexity. Those samples can be obtained by directly observing a
policy executed in the world.

Our goal is to first find a policy $\hat{\pi}$ such that 

\begin{align}
  \label{eq:1}
  & \hat{\pi} = \argmax_a \mathbb{E}_{s' \sim d_{\pi^{*}}}[s' \mid a,s],
\end{align}

and then to generate, at each iteration $i \in \{0, ..., N\}$, a new
policy $\tilde{\pi}_i$ that improves $\tilde{\pi}_{i-1}$, with
$\tilde{\pi}_0 = \hat{\pi}$. Such improvement is obtained by directly
executing $\tilde{\pi}_{i-1}$ and aggregating the reward measured over
several Monte Carlo simulations to the rewards at previous
iterations. Note that, (1) as in previous
work~\cite{Ross2011,Ross2014,Chang2015} we adopt a supervised learning
approach to imitate and learn a policy, (2) since the chosen actions
influence the distribution of states, our supervised learning problem
is characterized by a non-i.i.d.\footnote{Independent and identically
  distributed} dataset.

\subsection{Monte Carlo Search With Data Aggregation}
\label{sec:monte-carlo}

We now present Monte Carlo Search with Data Aggregation -- MCSDA, a
modification of the \textsc{AggreVate} and NRPI algorithms by Ross and
Bagnell~\cite{Ross2014} that (1) instead of the distribution of states
induced by the expert, always uses the learned policy to roll-in and
(2) rather than estimating the full cost-to-go of the policy, only
uses shorter Monte Carlo roll-outs to evaluate policy improvements.

In its simplest form the algorithm takes as input a set $\set{D}_e$ of
state-action pairs obtained from expert demonstrations and proceeds as
follows. First, MCSDA learns a classifier $\hat{\pi}$ by using
$\set{D}_e$ in order to imitate the expert. This is used to initialize
our policy $\tilde{\pi}$. Then, during each iteration, the algorithm
extends its dataset by (1) executing the previous policy $\tilde{\pi}$
and generating a state $s_t$ at each time-step, (2) selecting for each
$s_t$ an action $a_t$ that maximizes the expected value $V_p(s_t, a)$
of performing action $a$ at the given state, (3) aggregating the new
state-action pairs -- at each time-step -- to the previous
dataset. Finally, the aggregated dataset is used to train a new
classifier $\tilde{\pi}$ that substitutes the policy used at the
previous iteration. The details of MCSDA are provided in
Algorithm~\ref{alg:mcsda}.

\begin{algorithm}[t]
  \DontPrintSemicolon
  \SetAlgoLined
  \KwIn{$\set{D}_e$: dataset of state action pairs $\{s, a\}$ from 
    expert demonstrations, $N$: number of iterations of
    the algorithm, $K$: number of Monte Carlo simulations,
    $H$: simulation steps.}
  \KwOut{$\tilde{\pi}_N$: policy learned after N iterations of the
    algorithm.}
  \Begin{
    Train classifier  $\hat{\pi}$ on $\set{D}_e$ to imitate the expert.\;
    Set $\tilde{\pi}_{0} \leftarrow \hat{\pi}$.\;
    Initialize $\set{D} \leftarrow \set{D}_e$.\;
    \For{$i = 1$ \KwTo $N$}{
      Set $s_0$ in some state from the initial state distribution $D$.\;
      \For{$t = 1$ \KwTo $T$}{
        Get state $s_t$ by executing $\tilde{\pi}_{i-1}(s_{t-1})$.\;
        $\set{A} \leftarrow$ select or sub-sample (if needed) feasible actions in $s_t$.\;
        \ForEach{$a \in \set{A}$}{
          execute $K$ Monte Carlo simulations of length $H$ to
          estimate $V_p(s_t,a).$
        }
        Set $a_t \leftarrow \argmax_a V_p(s_t,a)$.\;
        Set $\set{D} \leftarrow \set{D} \cup \{s_t, a_t\}$.\;
      }

      Train classifier $\tilde{\pi}_{i}$ on $\set{D}$.\;
    }
    \BlankLine
    \Return{$\tilde{\pi}_{N}$}\;
  }
  \caption{Monte Carlo Search with Data Aggregation (MCSDA).}
  \label{alg:mcsda}
\end{algorithm}

By relying on data aggregation, MCSDA generates a sequence
$\tilde{\pi}_{1}, \tilde{\pi}_{2}, ..., \tilde{\pi}_{N}$ of policies
and preserves the main characteristics of algorithms like
\textsc{AggreVate} -- i.e., (1) it builds its dataset by exploring the
states that the policy will probably encounter during its execution,
(2) it can be interpreted as a Follow-The-Leader algorithm that tries
to learn a good classifier over all previous data and (3) can be
easily transformed to use an online learner by simply using the
dataset in sequence. However, the implementation of MCSDA is more
practical due to the reduced amounts of roll-outs generated from the
Monte Carlo simulation. Additionally, our algorithm always performs
the roll-in and the roll-out -- after the one-step deviation -- with
the learned policy. Still, it is worth to notice that the learned
policy is effectively generated from the mixture of sub-optimal expert
policies and learner's experience from Monte Carlo simulations. Hence,
expert policy actions will be likely executed at the beginning, while
their execution probability will reduce with the number of iterations
of the algorithm. This can be interpreted as combining the sub-optimal
policy of the expert and the learned policy with a varying mixing
parameter $\beta$ that initially is equal to $1$ -- always uses the
expert -- and decreases over subsequent iterations of
MCSDA. Consequently, we can rely on performances analogous to those
presented by Chang et al.~\cite{Chang2015}.

\subsection{Using MCSDA to Improve Robot Soccer Policies}
\label{sec:mcsda-improve-robot}

The application of MCSDA to the RoboCup context is not
straightforward, but requires an additional modeling effort. First, in
order to reduce the size of the problem, only the two-dimensional
position $p_r = (x_r, y_r)$ of the robot, the position $p_b = (x_b,
y_b)$ and velocity $v_b = (v_{xb}, v_{yb})$ of the ball in the field
have been adopted to represent the game state and, hence, to build our
learning dataset. Additionally, we generated the state-action pairs by
considering the following subset of actions: $\texttt{stand}$ (the
robot does not move), $\texttt{move\_up}$ (the robot moves forwards),
$\texttt{move\_down}$ (the robot moves backwards),
$\texttt{move\_left}$, $\texttt{move\_right}$.

\begin{figure}[t]
\centering
\includegraphics[width=\textwidth]{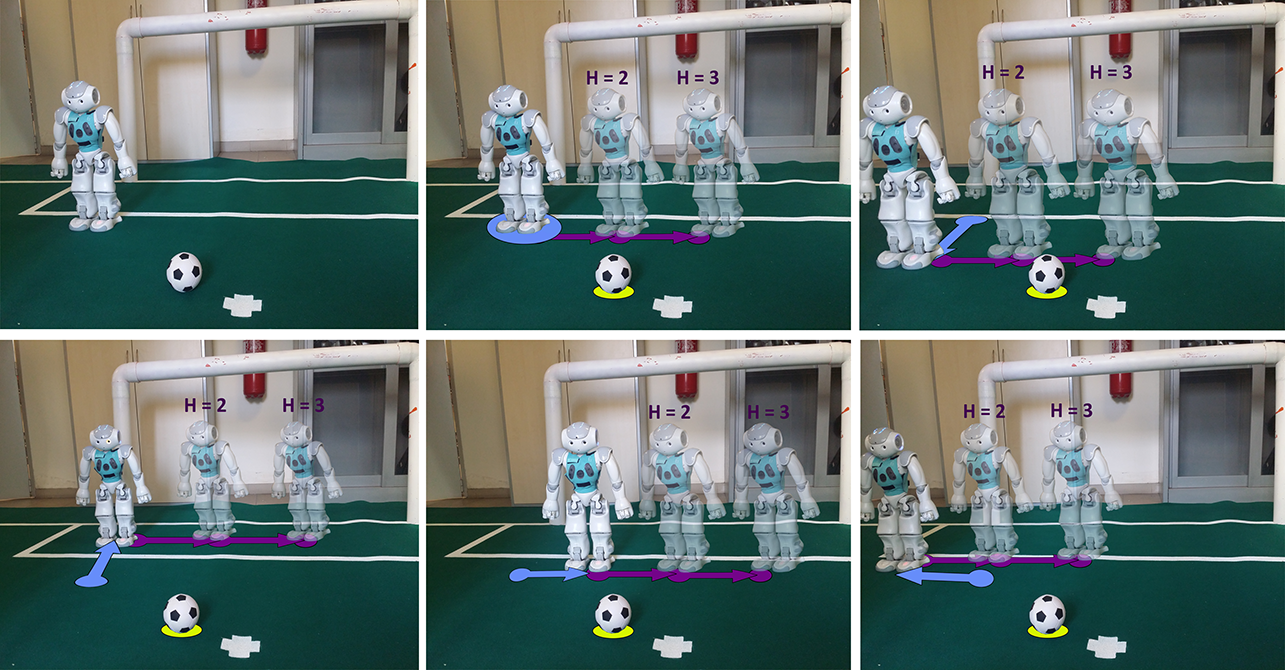}
\caption{Example of a full iteration of the Monte Carlo roll-outs: the
  robot evaluates all its actions, and selects the best one to
  maximize $V_p(s_t,a)$. In this example, the top-left sub-figure
  shows the world state at a given time $t$, and the current policy
  suggests the robot to execute $\texttt{move\_left}$. Accordingly,
  the other sub-figures show the evolution of the world state after
  each roll-out extending the current policy until the horizon $H =
  3$. The robot evaluates all the 5 actions: $\texttt{stand}$
  (top-center), $\texttt{move\_up}$ (top-right); $\texttt{move\_down}$
  (bottom-left); $\texttt{move\_left}$ (bottom-center);
  $\texttt{move\_right}$ (bottom-right). In these figures, the blue
  arrow represents the chosen action for the current roll-out, while
  the purple arrows represent the movements of the robot according to
  the current policy. The yellow circle represents the point $p_b$
  used to compute the reward according to Eq.~\ref{eq:reward}.}
\label{fig:actions_img}
\end{figure}

Given this reduced domain representation, as well as the goal of
generating a robot policy that adapts to the game adversaries, we
applied MCSDA to the RoboCup scenario by using the opponent team as
our expert. To this end, first we created a simple heuristic-based
classifier to recognize the opponents' actions with respect to the
ball (i.e., we collected their policy) and, then, learned such policy
in order to perform imitation. Note that at execution time the learned
policy is mapped to our robots by considering their relative position
with respect to the ball. In this work, such a mapping has been
manually defined. This resolves situations where the opponent (expert)
robot and our (learner) agent face the ball from opposite directions
and, for example, the opponent's $\mathtt{move\_left}$ action maps to
$\mathtt{move\_right}$ on our robot. Finally, Monte Carlo roll-outs
have been executed as illustrated in \figurename~\ref{fig:actions_img}
and using a reward function shaped as:

\begin{align}
  \label{eq:reward}
  R(s) = \frac{\mathtt{MAX\_FIELD\_DISTANCE} - |p_r - p_b|}{\mathtt{MAX\_FIELD\_DISTANCE}},
\end{align}

where $\mathtt{MAX\_FIELD\_DISTANCE}$ corresponds to the game-field
diagonal. To run our Monte Carlo simulations, we used both a
simplified simulator and a more complex one, provided by the B-Human
RoboCup Team\footnote{\url{https://www.b-human.de/}}.

\section{EXPERIMENTAL EVALUATION}
\label{sec:exper-eval}
RoboCup is a dynamic adversarial environment where robots needs to
adapt to the surroundings quickly and efficiently.  For these reasons,
the goal of this experimental section is to evaluate our learning
approach in the short range after few number of simulation steps. The
evaluation has been carried out through the B-Human soccer simulator
entirely written in C++ with the middle-sized humanoid NAO robot. In
this section, we test and validate the effectiveness of the two main
phases of our approach: the continuous policy improvement via Monte
Carlo roll-outs and the policy initialization via imitation
learning. In our experiments we set the roll-out horizon $H = 3$. This
value has been found to be a good trade-off between in-game
performance improvement and usability of the approach. Extending the
horizon, in fact, improves the player performance at the cost of more
computational resources.

\subsection{Policy Improvement}
\begin{figure}[t]
\centering
\includegraphics[scale=0.45]{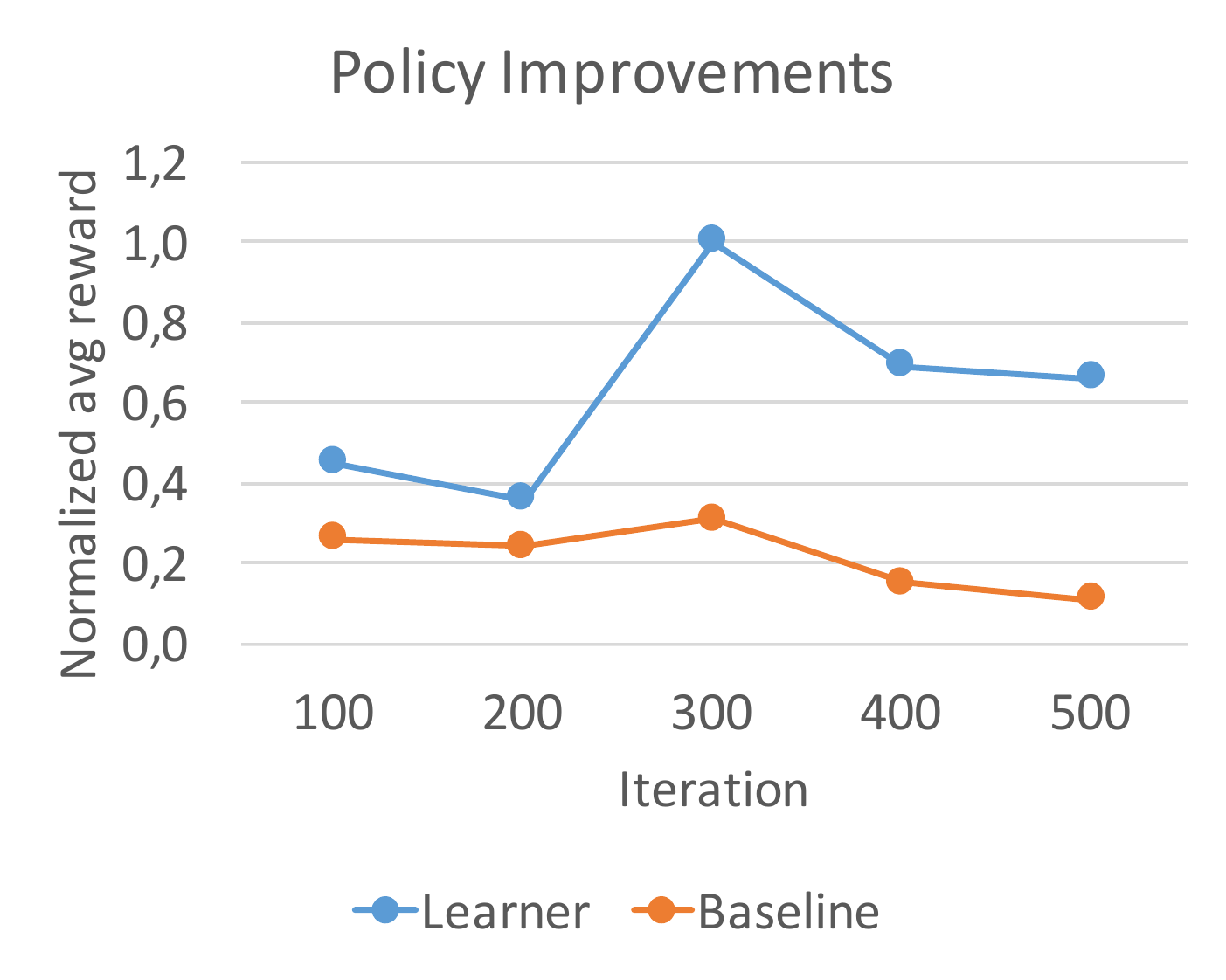}
\caption{Normalized average reward of the learner (\textit{blue}) and
  baseline (\textit{orange}) after different MCSDA iterations.}
\label{fig:opt_reward}
\end{figure}

The goal of our learner is to improve its performance while playing
against opponent robots and to decrease the number of opponent scores
while intercepting as many balls as possible. According to
Eq.~\ref{eq:reward}, we can evaluate each action of our learner by
considering the reward that the robot obtains during a match. Such a
measurement expresses how good the learner is positioned within the
field with respect to the ball. Therefore, we analyze the average
reward of our agent as well as the number of ball interceptions and
the final score of each match.  Fig.~\ref{fig:opt_reward} reports the
normalized average reward obtained by the learner during five regular
games, after a different number of MCSDA iterations. On the y-axis is
reported the obtained average reward.

Specifically, the learning defender features our MCSDA algorithm,
while the non-learning defender has a fixed policy initialized at
iteration zero. Such a baseline is a suitable comparison that allows
us to quantify the improvements of our robot in terms of positioning
with respect to its own initial policy. It is worth noticing that each
reported match has been played with different policies generated at
different iterations of MCSDA. Hence, each match represents a
different configuration of the learner, where its actions are
determined by a policy computed after 100, 200, 300, 400, 500
iterations of our algorithm. The plot shows a constant improvement
with respect to our baseline and over previous configuration of its
trained policy. It is worth remarking that the drop in performance
between game $3$ and $4$ can be due to different factors affecting the
game, such as player penalization and ball positioning rules. However,
such drop has a marginal impact with respect to the previous
improvements, and the performance of consequent matches remains
constant.

\begin{figure}[t]
\centering
\includegraphics[scale=0.45]{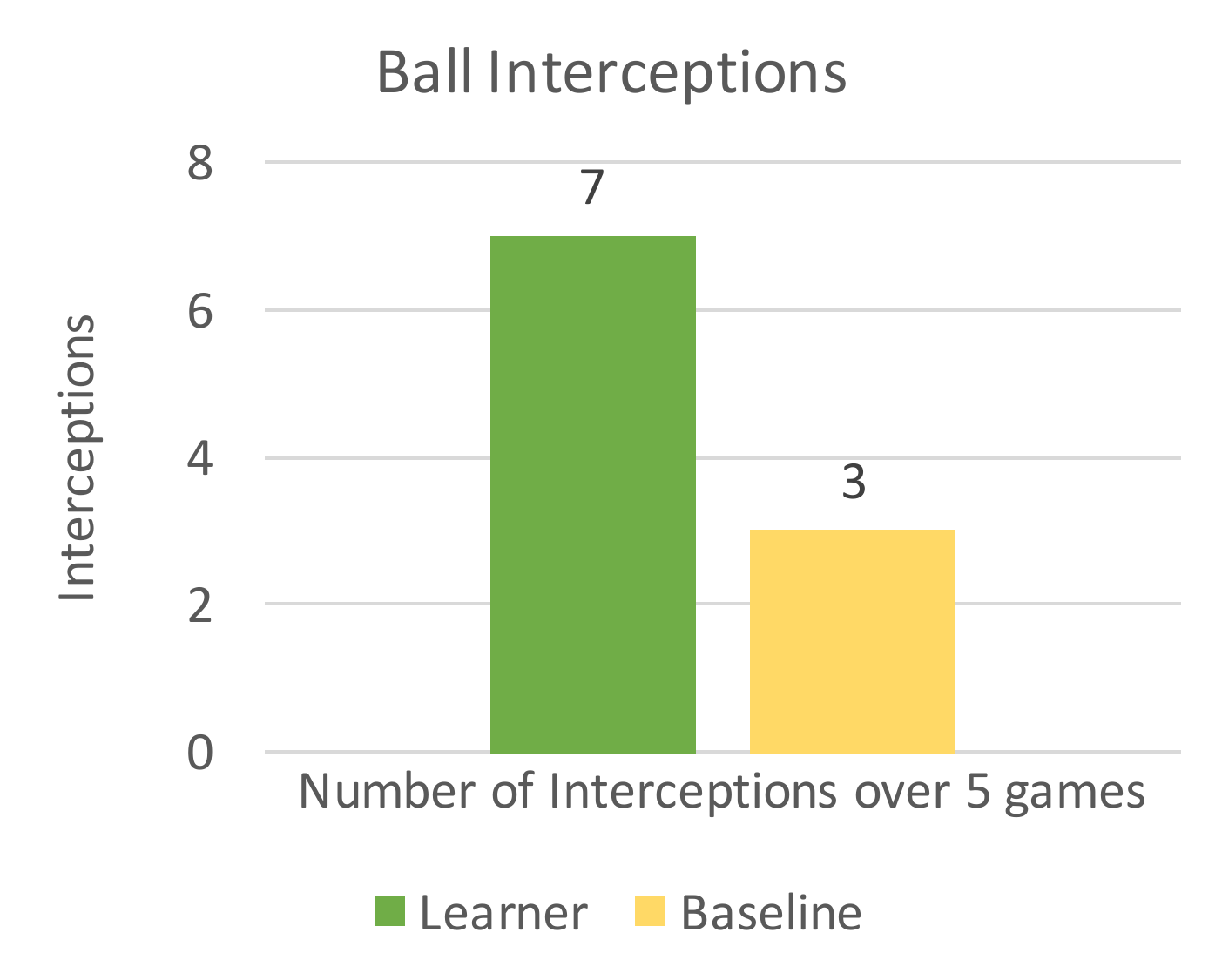}
\caption{Sum of intercepted ball over five matches after different
  MCSDA iterations.}
\label{fig:opt_recovers}
\end{figure}

Additionally, thanks to the nature of our testing environment, we are
able to report more direct evaluation indices for our approach. To
this end, we report the number of intercepted balls and the number of
opponent scores. In particular, \figurename~\ref{fig:opt_recovers}
shows the sum of intercepted balls of the two teams (learning and
non-learning) on the same set of games as before, and
Table~\ref{tab:opt_scores} reports their final scores.

\begin{table}[t]
\centering
\caption{The table reports the final scores of five matches after
  different MCSDA iterations.}
\label{tab:opt_scores}
\begin{tabular}{c||c|c|c|c|c|}
  & \multicolumn{5}{c|}{\textbf{MCSDA iterations}}\\\hline
  \textbf{Teams}~~ & \quad 100 ~~~  & \quad {200}  ~~~ & \quad {300}  ~~~ & \quad {400}  ~~~ & \quad {500}  ~~~ \\ \hline\hline
\quad Learning ~~~      & 2            & 3            & 0            & 1            & 1            \\ \hline
\quad Non-learning ~~~      & 3            & 2            & 1            & 1            & 1            \\
\end{tabular}
\end{table}

It is worth noticing that the number of intercepted balls of our
learning agent (\textit{green}) is more than twice the number of the
opponent defender (\textit{yellow}). Furthermore, the final results of
the different matches promises an interesting profile: even though the
learner does not win all of the matches, the number of opponent scores
decreases as the learner refines its policy. Since MCSDA is applied
only on defense robots, we do not achieve any improvement on the
number of goals of our team. However, as expected, by increasing the
number of iterations of our algorithm, the number of goals of the
opponent team decreases.
 
\subsection{Imitation Influence}

Since our robots operate in dynamic environments the policy training
process cannot be too long. Therefore, we need to restrict the search
space for our learning process as much as possible. To this end, we
generate an initial policy by running 100 matches with the only
purpose of analyzing most probable positions and velocities of the
ball, as well as opponents' positions within the field as introduced
in Section~\ref{sec:mcsda-improve-robot}.

\begin{figure}[t]
\centering
\includegraphics[scale=0.45]{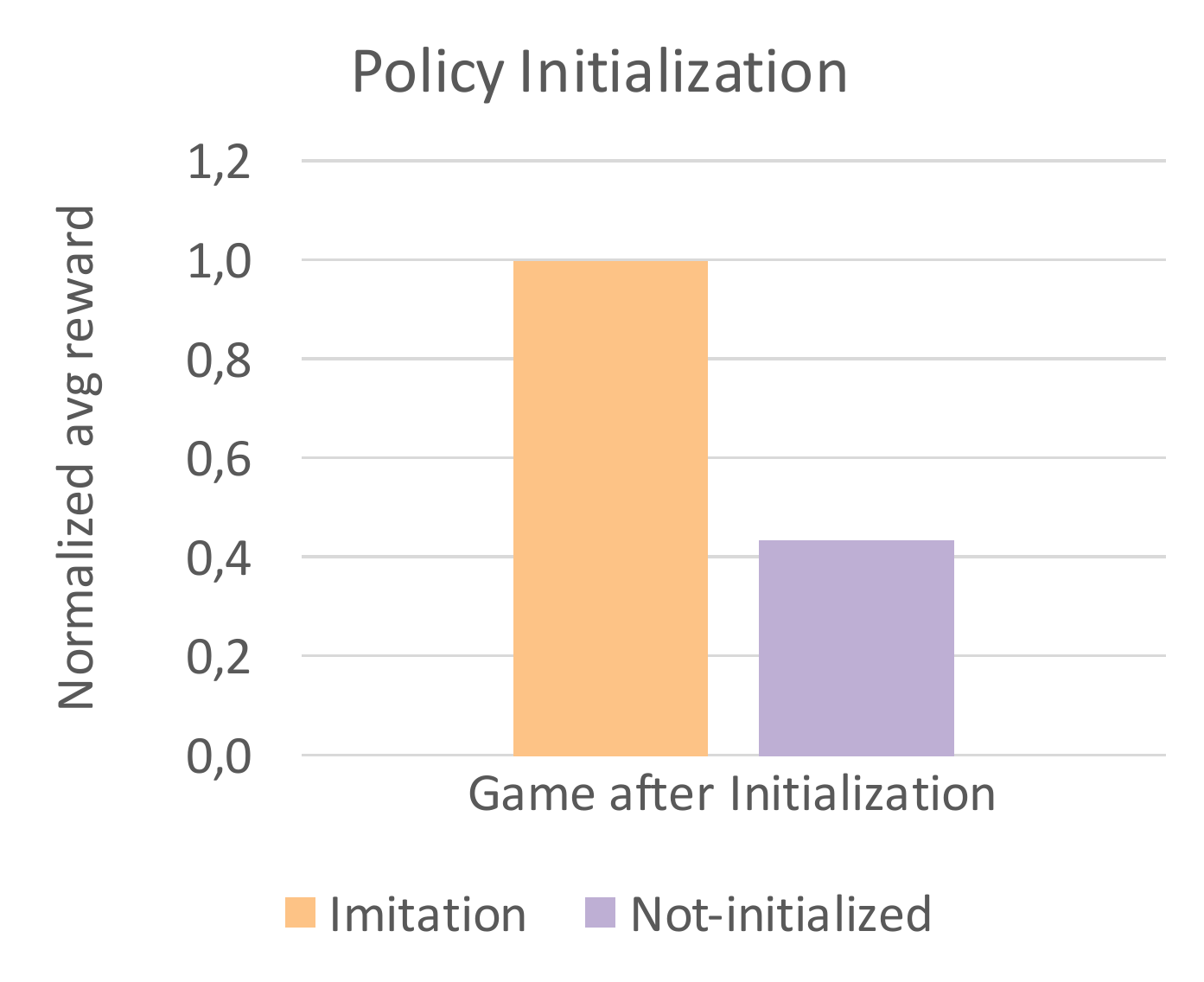}
\caption{Normalized average reward of the learner
  (\textit{orange}) and the non-initialized policy learner
  (\textit{purple}) during a match.}
\label{fig:im_reward}
\end{figure}

In this case, we setup an experimental evaluation with the aim of
studying the influence of our policy initialization on the overall
MCSDA approach. In the setting shown in
\figurename~\ref{fig:im_reward}, the blue team deploys a robot learner
featuring an initialized policy, while the red team deploys a learner
with a non-initialized policy. In this test, we let the two defenders
train their policies for 300 iterations. Afterwards, we select the two
different policy profiles in order to play a regular
match. \figurename~\ref{fig:im_reward} shows the normalized averaged
reward of the two learners: orange for the initialized policy, and
purple for the not-initialized one. It is worth noticing that -- as
expected -- an initialized policy significantly improves the learning
process.

\section{DISCUSSION AND FUTURE WORK}
\label{sec:conclusions}

In this paper we presented MCSDA, an algorithm that strictly relates
to recently developed approaches for policy improvement. We used and
evaluated our method to generate better strategies for soccer defense
in the RoboCup scenario. The application of MCSDA on this context
allowed our robots to increase the number of ball interceptions, as
well as to reduce the number of opponents' goals. Together with a
better performance, an overall more efficient positioning of the
defender player within the field has been achieved.

\subsection*{Contributions}
\label{sec:contributions}
The main contribution of this paper consists in the combination of
data aggregation together with Monte Carlo search. The use of Monte
Carlo search results in a practical algorithm, that allows a
real-world implementation on a robot domain. By relying on data
aggregation, instead, MCSDA preserves the main characteristics of
algorithms like \textsc{AggreVate} and can be easily transformed to an
online method. Finally, we also show that using MCSDA with a
simplified representation of the domain a good policy improvement can
be obtained on complex and challenging robotic scenarios.

\subsection*{Limitations and Future Work}
\label{sec:limitations}
Our algorithm still presents some limitations. In fact, even if Monte
Carlo simulations make MCSDA practical, it requires expensive calls to
a simulator and, hence, it has been applied to a single robot
player. While simple simulators can be used, neither the online
application of the algorithm, nor its use on a larger number of robots
are straightforward. For this reason, as future work, we would like to
adapt and test our algorithm to learn policies online and consequently
apply the algorithm to the whole robot soccer team. Furthermore, we
plan to perform additional studies on the performance guarantees that
MCSDA can achieve.

\bibliographystyle{splncs03}
\bibliography{bibliography}

\end{document}